\documentclass[conference]{IEEEtran}

\usepackage[utf8]{inputenc}
\usepackage[T1]{fontenc}
\usepackage{amsmath, amssymb, amsfonts}
\usepackage{graphicx}
\usepackage{subcaption}
\usepackage[font=footnotesize]{caption}
\usepackage{booktabs, array, tabularx, multirow}
\usepackage{float}
\usepackage{enumitem}
\usepackage[colorlinks=true, allcolors=blue]{hyperref}
\usepackage{stfloats}
\usepackage[ruled, linesnumbered]{algorithm2e}
\usepackage{cite}
\usepackage{balance}

\usepackage{tikz}
\usepackage{eso-pic}

\renewcommand{\baselinestretch}{0.952}
\newcommand\PlaceConfHeaderAndCopyright{
  \AddToShipoutPictureFG*{
    \put(50,10){
      \parbox[c]{6.5in}{
        
      }
    }
  }
}
\PlaceConfHeaderAndCopyright

\begin{document}

\title{Explainable AI for Mental Health Prediction in Drug-Affected Populations with Dragonfly Algorithm and GAN Oversampling}

\author{\IEEEauthorblockN{Ahnaf Atef Choudhury}
\IEEEauthorblockA{\textit{Department of Information Sciences and Technology} \\
\textit{George Mason University}\\
USA \\
achoudh9@gmu.edu}
\and
\IEEEauthorblockN{Shahriar Siddique Ayon}
\IEEEauthorblockA{\textit{Department of Computer Science} \\
\textit{American International University-Bangladesh}\\
Bangladesh \\
shahriarayon63@gmail.com}
\and
\IEEEauthorblockN{Md. Ebrahim Hossain}
\IEEEauthorblockA{\textit{Department of Computer Science} \\
\textit{American International University-Bangladesh}\\
Bangladesh \\
sbebrahim999@gmail.com}

\and
%\IEEEauthorblockN{Md. Parvej Hoque Palash}
%\IEEEauthorblockA{\textit{Department of Computer Science and Engineering} \\
%\textit{Jahangirnagar University}\\
%Bangladesh \\
%palash.stu2018@juniv.edu}
%\and

%\IEEEauthorblockN{Ramkrishna Saha}
%\IEEEauthorblockA{\textit{Department of Computer Science} \\
%\textit{The University of Texas at Dallas}\\
%USA \\
%ramkrishna.saha@utdallas.edu}
%\and
\IEEEauthorblockN{Abdullah Al Mamun}
\IEEEauthorblockA{\textit{Department of Computer Science and Engineering} \\
\textit{Dhaka University of Engineering and Technology}\\
Bangladesh \\
mamun.duet.bd@gmail.com}
}

\maketitle

\begin{abstract}
Mental illnesses among drug users are an increasing international issue, particularly in regions where early detection cannot be easily undertaken. The current literature tends to ignore the use of AI-based mental health analysis in drug users, and low quality of the class imbalance treatment, low interpretability, and optimal hyperparameter optimization can lower predictive quality and clinical utility. This study present a detailed, explainable machine learning (ML) model of multiclass mental health prediction, using a multidimensional data set of drug-affected persons. We combine hybrid PCA-Information Gain (PCA-IG) feature selection, Generative Adversarial Network (GAN)-based oversampling, and Dragonfly Algorithm (DA)-optimized XGBoost to address some of the limitations of existing methods. The suggested framework is effective to work with high-dimensional categorical data, address the issue of class imbalance, and improve predictive performance due to intelligent hyperparameter tuning. The experimental findings show that the XGBoost model optimized using the DA, in combination with GAN-based oversampling, has an accuracy of 94.17\% and a weighted F1-score of 93.80\%, which is better than the traditional and baseline models. The behavioral, lifestyle, and health factors, particularly sleep quality, physical health, and emotional regulation, are strongly predictive of mental health, with demographic factors having little impact, as seen through feature analysis. SHAP-based explainable AI provides easy-to-understand, instance-level information, enhancing interpretability and trust in models to be used in clinical settings. The results indicate that this framework has the potential to generate valid mental health forecasting tools, which would facilitate early intervention and enhance the treatment of drug-influenced people.
\end{abstract}


\section{Introduction}
Mental health and substance use disorders (SUD) are an increasingly significant health problem in the world that tends to co-occur and strengthen one another. Drug addicts are also prone to increased chances of anxiety, depression, and emotional instability, which may perpetuate substance use and hamper recovery \cite{AIHW2025}. Mental illnesses and SUD are significant contributors to the burden of disease in the world, having been ranked within the top ten causes of illness in the world \cite{Erskine2015}. Approximately 20\% of children and adolescents worldwide suffer from a mental disorder \cite{UNICEF2024}. In 2024, the World Health Organization (WHO) listed 2.6 million deaths each year related to alcohol (4.7 percent of all deaths) and 0.6 million related to the use of psychoactive drugs \cite{PAHO2024}. Epidemiology indicates that the abuse of illegal drugs is common with severe consequences to the health of the individual and health care systems. Proper mental health risk discrimination is essential in resource-limited environments such as Bangladesh, where drug addiction has intersected social, behavioral, and cultural issues, since early intervention and targeted support are imperative \cite{Appleton2025,Ayon_10.1371}.

The conventional diagnostic techniques use subjective, time-intensive tests and frequently overlook subtle trends, which limit their validity and scalability \cite{Ayon_Advancing_2025}. Machine learning (ML) methods have become promising alternatives, and they have shown promising outcomes predicting mental health outcomes in psychoactive drug users \cite{Toni2024}. Bari et al. examined 3,476 adults based on 15 judgment and contextual variables to predict SUD behaviors, recency of substance use, and severity with an 83\% accuracy and AUC of 0.74 \cite{Bari2026}. Some of the ML models used by Ndikumana et al. include Random Forest, which had the highest accuracy of 88.8\% \cite{Ndikumana2025}. However, most existing models suffer from three major limitations: (1) inadequate handling of imbalanced data where critical “Poor” cases are underrepresented; (2) limited transparency that makes interpretation difficult for clinicians; (3) suboptimal parameter tuning that reduces performance on high-dimensional, categorical health data.

Advanced analytics and AI have been leveraged to develop decision support systems (DSS) for more accurate mental health diagnosis \cite{Tutun2023, abubakkar2025explainable}. Imbalance in class data negatively influences the performance of typical ML algorithms. To tackle existing challenges, feature ranking is performed using IG, dimensionality is reduced through PCA, and explainable AI (XAI) techniques are incorporated to enhance transparency. Additionally, a swarm-intelligence-based optimization strategy enhances model generalization, while Generative Adversarial Networks (GAN)-based oversampling effectively addresses class imbalance. SHAP is used to explain individual predictions, showing that sleep, emotional regulation, physical health, and substance use strongly impact mental health outcomes. Key Contributions of this work:

\begin{itemize}
    \item First integration of Dragonfly Algorithm (DA) with XGBoost for mental health prediction in drug-affected populations.
    \item Hybrid PCA-IG feature selection that balances dimensionality and predictive power with sleep quality, access to healthcare, and behavioural factors as the most important.
    \item GAN-based multi-class mental health classification oversampling, which is better than SMOTE and SVM-SMOTE.
    \item SHAP-based explainability with actionable clinical insights using waterfall and force plots.
\end{itemize}

The rest of the paper is structured as follows: Sections \ref{sec:Lr}, \ref{sec:Method}, \ref{sec:Result}, and \ref{sec:Con} present the related work, methodology, results, and conclusion, respectively.

\section{Literature Review}
\label{sec:Lr}
In recent years, mental health disorders have become more and more a subject of attention because of their complicated interdependence and the increasing burden on the world. Current research has been using the ML method to forecast mental health risks in drug-impacted groups. Brito et al. \cite{Brito_10068843} showed that Support Vector Machines (SVM) and Sequential Backward Selection (SBS) showed high predictive accuracy of common mental disorders (CMD) and depression in psychoactive drug users with accuracies of 82.81\% and 81.98\% respectively. In a similar study, Zhang et al. \cite{Zhang_11167906} also used logistic regression, SVM, and random forests to predict results of drug consumption; their results show that SVM performed the best (AUC = 0.7821, F1 = 0.6608).  

Acharya et al. improved the predictive modeling by using the combination of Random Forest and XGBoost, which proved to be effective in identifying co-occurring mental health and substance use disorders in women (AUC = 0.817, Accuracy = 0.751) \cite{Acharya_14041630}. Similar results were also described by Yifan et al. \cite{Yifan_10213032}, the Cross Gradient Booster (Random Forest) model was the most accurate (0.79784) at predicting mental health outcomes based on survey data. Epidemiological data highlight the magnitude of the problem with 59\% of Americans over 12 years and above indicating that they have ever used illegal drugs illegally \cite{ErnestOkonofua2025}. Zaha et al. examined 203 patients with dual diagnosis, with cannabis, novel psychoactive substances, and unknown drugs being the most commonly used substances \cite{Zaha_13192543}. These results underline the significance of considering demographic and clinical characteristics in predictive models.

The strategies of feature selection and class imbalance have improved the model reliability and scalability. According to Chigagure et al., such methods are significant to enhance predictive robustness \cite{chigagure2025machine}. GANs have especially been useful in solving class imbalance, through creation of synthetic data, which enhances prediction accuracy \cite{Zhang2025, aziz2025comprehensive}. Narteni et al. generalized this use to the high-dimensional, low-sample-size data, proving the helpfulness of GANs in data extrapolation to small datasets \cite{NARTENI2025110133}. In addition to classical ML, new algorithms have been considered. Wrapper-based feature selection has demonstrated potential in the classification of illnesses by the DA \cite{Sekhar2024}. These bio-inspired optimization techniques underscore the possibility of AI helping to optimize predictive models of complex health outcomes. Table~\ref{tab:comparison} summarizes recent studies on mental health and drug use prediction, comparing datasets, methods, accuracy, XAI usage, and feature engineering with our proposed framework.

\begin{table*}[htbp]
\centering
\caption{Comparison of Related Works on Mental Health and Drug Use Prediction}
\label{tab:comparison}
\footnotesize
\renewcommand{\arraystretch}{1.15}
\setlength{\tabcolsep}{4pt}
\begin{tabular}{c p{3.5cm} p{3.5cm} p{2.6cm} c p{2.8cm}}
\toprule
\textbf{Ref.} & \textbf{Dataset} & \textbf{Method} & \textbf{Accuracy} & \textbf{XAI} & \textbf{Feature Engineering} \\
\midrule
%\cite{Brito_10068843} & Psychoactive drug users survey & SVM + Sequential Backward Selection (SBS) & 82.81\% & No & SBS-based selection \\
%\cite{Zhang_11167906} & Drug consumption dataset & Logistic Regression, SVM, Random Forest & 78.21\% & No & Basic preprocessing \\
\cite{Acharya_14041630} & Women with co-occurring disorders & Random Forest + XGBoost & 75.10\% & No & Demographic \& clinical features \\
\cite{Yifan_10213032} & Mental health survey data & Cross Gradient Booster (Random Forest) & 79.78\% & No & Survey-based features \\
%\cite{Zaha_13192543} & 203 dual-diagnosis patients & Statistical / clinical analysis & 72.40\% & No & Clinical profiling \\
%\cite{chigagure2025machine} & Mental health dataset & ML with feature selection \& class balancing & 81.50\% & No & FS, imbalance handling \\
\cite{Zhang2025} & Imbalanced health dataset & GAN-based synthetic data + ML & 85.30\% & No & GAN augmentation \\
%\cite{NARTENI2025110133} & High-dim, low-sample dataset & GAN data extrapolation & 83.60\% & No & GAN augmentation \\
\cite{Sekhar2024} & Disease classification dataset & Dragonfly Algorithm (wrapper FS) & 86.70\% & No & Bio-inspired wrapper FS \\
\cite{Squires2023} & Multimodal (clinical, neuroimaging, behavioral) & Deep learning / multimodal AI & 88.20\% & Partial & Multimodal fusion \\
\cite{Humayun2025} & Mental health dataset & ML with SHAP \& LIME & 89.40\% & Yes (SHAP, LIME) & Standard preprocessing \\
\midrule
\textbf{Proposed} & Drug-affected individuals (multidimensional)  & \textbf{PCA-IG + GAN oversampling + DA-XGBoost} & \textbf{94.17\%} & \textbf{Yes (SHAP)} & \textbf{PCA-IG; behavioral, lifestyle \& health} \\
\bottomrule
\end{tabular}
\end{table*}

Current research on mental health predicting in drug-affected populations is constrained by data imbalance, biased modelling, and insufficient interpretability, frequently overlooking essential behavioural and psychological factors. This work deals with these problems by using a balanced and optimised framework that combines hybrid PCA-IG feature selection, GAN-based oversampling, DA-optimized XGBoost, and SHAP-driven explanation.

\section{Methodology}
\label{sec:Method}
\begin{figure}[htbp]
    \centering
    \includegraphics[width=0.45\textwidth]{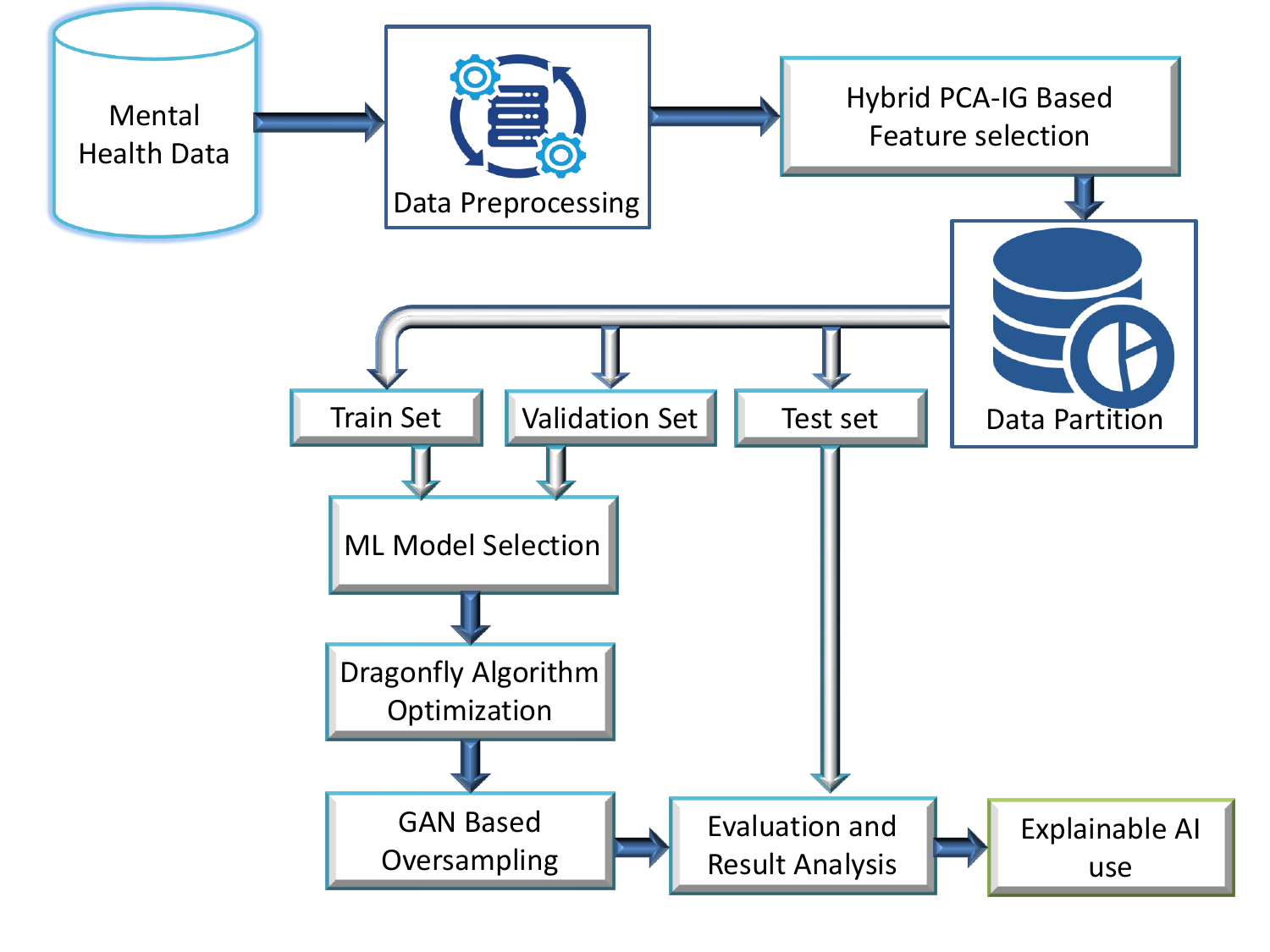}
    \caption{Proposed Framework for Mental Health Risk Prediction in Drug-Affected Populations.}
    \label{fig:method}
    \end{figure}

In this section, the proposed scheme is described, such as data collection and preprocessing, feature importance analysis, model selection with hyperparameter optimization, the issue of class imbalance through oversampling, and interpretability through XAI. The entire process of risk prediction of mental health in drug-influenced groups is depicted in Figure \ref{fig:method} .

\subsection{Data Collection and Preprocessing}
The dataset of the present paper, 'Insights into Drug Addiction in Bangladesh: A Multidimensional Dataset', will provide a closer perspective of the variables associated with drug use and mental health in Bangladesh \cite{Islam2024}. It consists of 36 features (categorical) that include personal data like age, sex, and religion, as well as socioeconomic factors like education, employment, and economic status. Moreover, the dataset reflects behavioral patterns (e.g. smoking, social activities and crime) and significant psychological and health related variables, such as emotional issues, sleep quality, dietary habits, trauma history, and suicidal ideation. It also includes drug-use behavior in terms of type of substance, age of first use, frequency of use and the effect of friends and family. The target variable, Mental\_Health\_Status, will be categorical, consisting of three classes (Good, Average, and Poor) and the problem may be defined as a multi-class classification problem. 

Upon acquisition, the dataset underwent systematic preprocessing to ensure completeness, consistency, and analytical readiness. All unnecessary columns were removed, and there were no missing values. All features were categorical and thus we used label encoding to turn them into numbers. In the case of the target variable, Mental\_Health\_Status, the three classes were coded to be: 0= Average, 1= Good, 2= Poor. Having done all these, the dataset was now ready to undergo classification analysis.

\subsection{Feature Contribution and Selection}

\begin{figure*}[h]
    \centering
    \includegraphics[width=\linewidth]{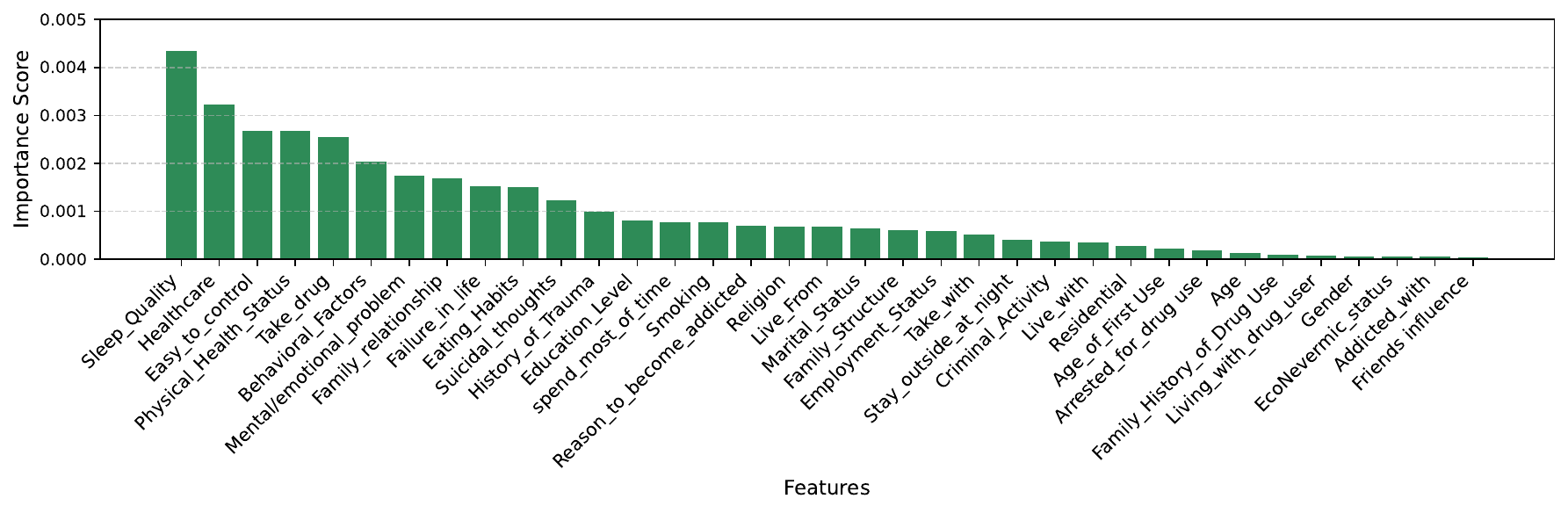}
    \caption{Ranking of Features Using Hybrid PCA–IG for Mental Health Prediction.}
    \label{fig:Feature}
    \end{figure*}

Following preprocessing, we experimented with various feature selection techniques and discovered that Principal Component Analysis-Information Gain (PCA-IG) worked best. This approach, which builds on earlier studies, combines PCA for dimensionality reduction with Information Gain to choose the most pertinent features, improving the overall performance of the model \cite{omuya2021feature1}. The ranking of feature importance of PCA-IG is shown in Figure \ref{fig:Feature}.

Figure \ref{fig:Feature} demonstrates the hybrid PCA-IG feature importance and their respective scores to predict mental health. The top score is Sleep\_Quality with 0.00435, then Healthcare (0.00322), and Easy\_to\_control (0.00268) where physical and mental health influence considerably. Physical\_Health\_Status (0.00267) and Take drug (0.00255) are crucial factors with behavioral and emotional factors such as Behavioral factors (0.00205), Mental/emotional problem (0.00175), Family relationship (0.00169) and Eating habits (0.00151) playing significant roles. Conversely, less influential are the demographic factors like Gender (0.000066) and Friends influence (0.000040) which mean that mental health in this dataset is more health and behaviour based than demographic.

After reduction of features, we narrowed down on 16 important features in 1,215 instances. The data was split into 80\% training and 20\% test with 20\% of the training data being further split into validation to facilitate consistent and sound model performance.

\subsection{Hyperparameter Optimization and Model Selection}
After experimenting with a variety of models, we chose the most successful ones, and we used popular Python libraries to assess their performance in our mental health prediction problem. The traditional models such as SVR and KNN performed worse and were thus not included in the findings. A linear baseline model was a Logistic Regression (LR) having a regularization parameter C = 1.0. Random Forest Classifier (RFC) (200 trees, depth = 10), Gradient Boosting (GB) (150 estimators, LR = 0.1), and LightGBM (250 leaves, LR = 0.05) were configured, while XGBoost achieved best performance with 300 estimators and depth = 8. The Artificial Neural Network (ANN) used two hidden layers (64, 32 neurons) with ReLU activation and LR = 0.001. XGBoost was the best performing among these models and we also optimized it with the advanced swarm intelligence and evolutionary algorithms. We used Particle Swarm Optimization, Artificial Bee Colony, Ant Colony Optimization and Dragonfly Algorithms (DA) and the latter had the best predictive performance improvement. The DA is a swarm intelligence algorithm that is based on dragonfly swarming and is a balance between exploration and exploitation. All dragonfly agents are candidate solutions and the position of each agent is updated via social interaction and optimization dynamics as:

\begin{equation}
    \mathbf{X}_i^{t+1} = \mathbf{X}_i^t + s \mathbf{S}_i + a \mathbf{A}_i + c \mathbf{C}_i + f \mathbf{F}_i + e \mathbf{E}_i
\end{equation}

where $\mathbf{X}_i^t$ is the position of the $i$-th agent at iteration $t$, $\mathbf{S}_i$, $\mathbf{A}_i$, $\mathbf{C}_i$ are separation, alignment, and cohesion, $\mathbf{F}_i$ and $\mathbf{E}_i$ represent attraction to food and distraction from enemies, with $s, a, c, f, e$ as weighting factors.

The DA-optimized XGBoost with $\eta = 0.05$, $d = 8$, $n = 300$, $s = 0.8$, and col\_sample\_bytree $= 0.7$ enhanced generalization and prediction for multiclass mental health classification.

\subsection{Oversampling Strategies for Class Imbalance}
The target column, Mental\_Health\_Status, is imbalanced with 766 samples for Average, 248 for Good, and 201 for Poor. To address the class imbalance, we applied oversampling techniques exclusively to the training set to create a balanced distribution, while keeping the test set unchanged to ensure unbiased evaluation. Among the oversampling methods tested, Synthetic Minority Over-sampling Technique (SMOTE), SVM-SMOTE, and our proposed (GAN)-based approach performed best. GAN generate realistic synthetic samples for minority classes by training a generator $G$ to produce data that a discriminator $D$ cannot distinguish from real data. The objective is:

\begin{equation}
\begin{split}
    \min_G \max_D V(D, G) = \mathbb{E}_{x \sim p_{\text{data}}}[\log D(x)] + \mathbb{E}_{z \sim p_z}\\
    [\log(1 - D(G(z)))]
\end{split}
\end{equation}

where $x$ is a real sample, $z$ is random noise, $G(z)$ is the generated sample, and $D(\cdot)$ predicts real versus fake.

\subsection{Explainable AI Integration}
XAI seeks to make model decisions transparent \cite{ayon2024harvesting}. In this study, we employed SHAP to interpret predictions by measuring the contribution of each feature using game theory concepts. This relationship is formalized in Equation \ref{eqn:shap}.

{\small
\begin{equation}
\label{eqn:shap}
\begin{split}
\phi_{i}(x) = 
\sum_{S \subseteq \{1, \dots, P\} \setminus \{i\}}
\frac{|S|! \, (P - |S| - 1)!}{P!}
\bigl[
f_{i}(S \cup \{i\})
\\
- f_{x}(S)
\bigr].
\end{split}
\end{equation}
}

The SHAP value $(\phi_i)(x)$ quantifies how much feature $i$ contributes to a prediction $x$ by comparing the model’s output with and without that feature across all feature subsets, where $f_x(S)$ is the model’s prediction for subset $S$ of the total $p$ features.

\subsection{Evaluation Metrics}

For multiclass classification, weighted metrics were used to address class imbalance. Weighted precision, recall, and F1-score indicate class-specific performance proportionally, but total accuracy shows overall model correctness, ensuring a full evaluation across all mental health classes. 

%The metrics of evaluation are determined as:

% \begin{equation}
% \begin{aligned}
% \text{Acc} &= \tfrac{TP+TN}{TP+TN+FP+FN}, \quad
% P_w = \sum_{i=1}^{C} w_i \tfrac{TP_i}{TP_i+FP_i}, \\[4pt]
% R_w &= \sum_{i=1}^{C} w_i \tfrac{TP_i}{TP_i+FN_i}, \quad
% F1_w = \sum_{i=1}^{C} w_i \tfrac{2 P_i R_i}{P_i+R_i}
% \end{aligned}
% \end{equation}

% where $C$ is the number of classes and $w_i = n_i / N$ is the weight of class $i$ based on its sample proportion.

\section{Results Analysis and Discussion}
\label{sec:Result}
Our ML models were developed and tested on the free version of Google Colab, offering easy access and collaboration. To improve the accuracy and minimise overfitting, we used 5-fold cross-validation on all models. This part presents a summary of the prediction outcomes using various models, including the performance variations prior to and after feature selection and oversampling, and an easy-to-follow step-by-step account of our method.

\subsection{Initial ML Model Performance without Data Enhancement}
Table~\ref{tab:baseline_performance} shows model performance on multiclass mental health prediction before feature selection and oversampling. LR achieved the lowest accuracy at 70.81\%, while XGBoost performed best among standard models with 88.15\% accuracy. Ensemble methods like RFC and LGBM performed well, with accuracies of 87.26\% and 86.44\%, respectively, and the ANN achieved 83.18\% accuracy. Notably, the DA-optimized XGBoost further improved results, achieving the highest accuracy of 89.32\% and a weighted F1-score of 88.74\%, outperforming all other models.

\begin{table}[htbp]
\caption{Model Performance Before Feature Selection and Oversampling}
\label{tab:baseline_performance}
\centering
\resizebox{\columnwidth}{!}{
\begin{tabular}{p{1.7cm}p{1.1cm}p{1.1cm}p{1.1cm}p{1.cm}c}
\hline
\textbf{Model Name} & \textbf{Weighted Precision (\%)} & \textbf{Weighted Recall (\%)} & \textbf{Weighted F1-Score (\%)} & \textbf{Accuracy (\%)} \\ \hline
LR & 68.62 & 70.76 & 69.02 & 70.81 \\ \hline
LGBM & 86.24 & 86.42 & 86.01 & 86.44 \\ \hline
GB & 84.25 & 84.03 & 84.02 & 84.42 \\ \hline
ANN & 82.82 & 83.11 & 82.93 & 83.18 \\ \hline
RFC & 87.61 & 87.27 & 87.24 & 87.26 \\ \hline
XGBoost & 88.12 & 88.14 & 88.02 & 88.15 \\ \hline
DA-XGBoost & 89.22 & 88.28 & 88.74 & 89.32 \\ \hline
\end{tabular}
}
\end{table}

\subsection{Enhanced Model Performance with Feature Selection and Oversampling}

After applying feature selection and oversampling, all models showed significant improvement (Table~\ref{tab:performance_comparison}). Using SMOTE, LGBM achieved 88.23\% accuracy, RFC 90.08\%, and XGBoost 91.14\%. SVMSMOTE further boosted performance, with ANN reaching 88.47\%, XGBoost 91.28\%, and DA-optimized XGBoost 92.08\%. GAN-based oversampling produced the best results overall: RFC achieved 91.52\%, XGBoost 93.04\%, and DA-XGBoost led with 94.17\% accuracy and a weighted F1-score of 93.80\%, demonstrating the combined effectiveness of feature selection, advanced oversampling, and hyperparameter optimization.

\begin{table}[htbp]
\caption{Performance Comparison of Oversampling and ML Models}
\centering
\resizebox{\columnwidth}{!}{
\begin{tabular}{p{1.2cm} l  p{1.1cm} p{1.1cm} p{1.1cm} p{1cm}}
\hline
\textbf{Oversam-pling Method} & \textbf{ML Model} & \textbf{Weighted Precision (\%)} & \textbf{Weighted Recall (\%)} & \textbf{Weighted F1-Score (\%)} & \textbf{Accuracy (\%)} \\
\hline
     & LGBM        & 87.42 & 88.18 & 88.06 & 88.23 \\
SMOTE     & RFC         & 89.31 & 89.22 & 89.38 & 90.08 \\
    & XGBoost     & 90.08 & 89.74 & 89.82 & 91.14 \\ \hline
 SVM- & ANN         & 87.38 & 88.10 & 88.21 & 88.47 \\
SMOTE  & XGBoost     & 90.13 & 89.83 & 89.87 & 91.28 \\
  & DA-XGBoost  & 91.21 & 91.12 & 91.24 & 92.08 \\\hline
      & RFC         & 91.20 & 91.42 & 91.30 & 91.52 \\
GAN       & XGBoost     & 92.26 & 92.38 & 92.28 & 93.04 \\
      & DA-XGBoost  & 93.75 & 93.85 & 93.80 & 94.17 \\
\hline
\end{tabular}
}
\label{tab:performance_comparison}
\end{table}

\begin{figure}[htbp]
    \centering
    \includegraphics[width=\linewidth]{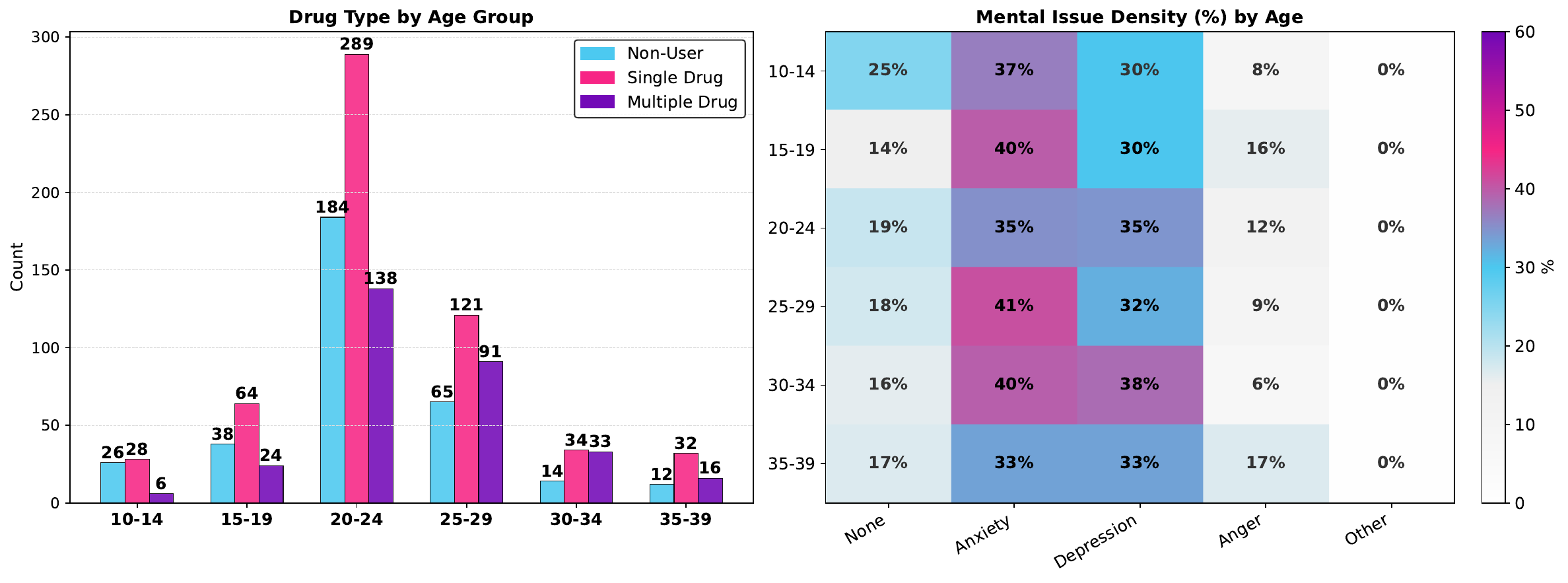}
    \caption{Dual Analysis of Age-Wise Drug Use and Mental Health Patterns.}
    \label{fig:Dual Analysis}
    \end{figure}

Figure \ref{fig:Dual Analysis} shows the dual perspective of the age-based pattern of drug use and mental health patterns with emphasis on their behavioral and psychological interdependence. The bar chart on the left indicates that there were the highest numbers of drug users within the 20-24 age group with 289 single drug users, 138 multiple drug users and 184 non-users and then moderate drug use in 25-29-year-olds (121 single, 91 multiple users). Early teenagers (1014) exhibit little exposure to multiple drugs (6). The right heatmap indicates that anxiety and depression prevail across age, with the highest of 41\%(25-29) and 38\%(30-34) respectively; the highest percentage of the 20-24 age group had both anxiety and depression (35\% each), coinciding with the maximum drug use. Anger levels are low (6-17\%), and other problems are not significant, with internalizing disorders being the primary mental health burden. The age range of 20-29 years is a critical period of overlap between peak drug use and high psychological distress, which is suitable as a target of interventions to minimize substance abuse and chronic psychological risk pathways.

\subsection{SHAP-Based Insights into Model Decisions}
\begin{figure}[htbp]
    \centering
    \includegraphics[width=\linewidth]{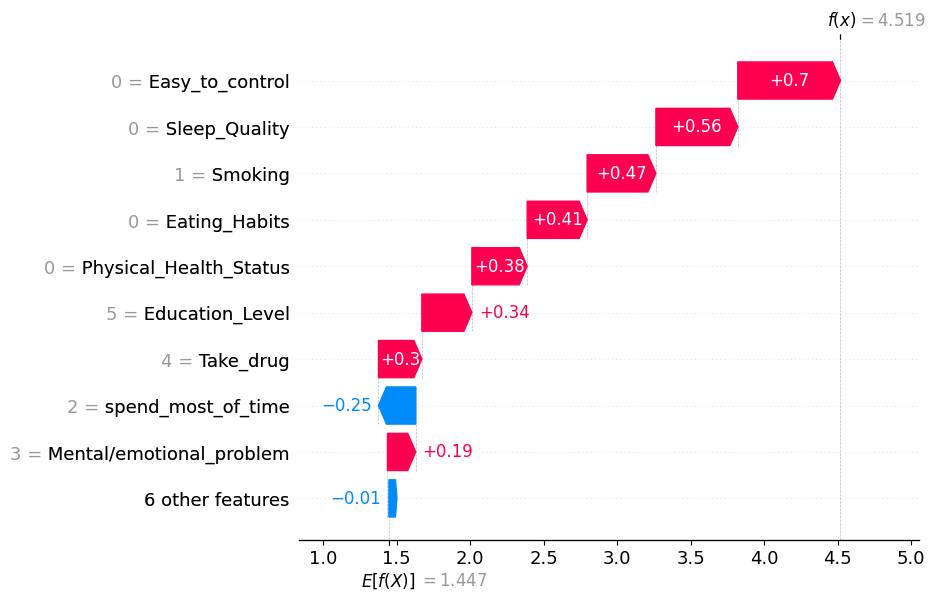}
    \caption{SHAP Waterfall Plot for Sample 10 Showing Feature Contributions.}
    \label{fig:SHAP1}
    \end{figure}

SHAP waterfall and force plots are used to visualize the effect of individual features on the classification predictions of the model in XAI. The SHAP waterfall plot \ref{fig:SHAP1} represents the SHAP waterfall model, with sample\_index = 10, with a baseline prediction $E[f(X)] = 1.447$ pushed to a final prediction of $f(x) = 4.519$, which is a high-risk class. The strongest positive contributors are Easy\_to\_control (+0.70), Sleep\_Quality (+0.56), and Smoking (+0.47), followed by Eating\_Habits (+0.41), Physical\_Health\_Status (+0.38), Education\_Level (+0.34), and Take\_drug (+0.30). Minor effects include spend\_most\_of\_time ($-0.25$), Mental/emotional\_problem (+0.19), and others ($\sim -0.01$). On the whole, behavioral and lifestyle aspects prevail in the decision of the model on this case.

\begin{figure*}[htbp]
    \centering
    \includegraphics[width=\linewidth]{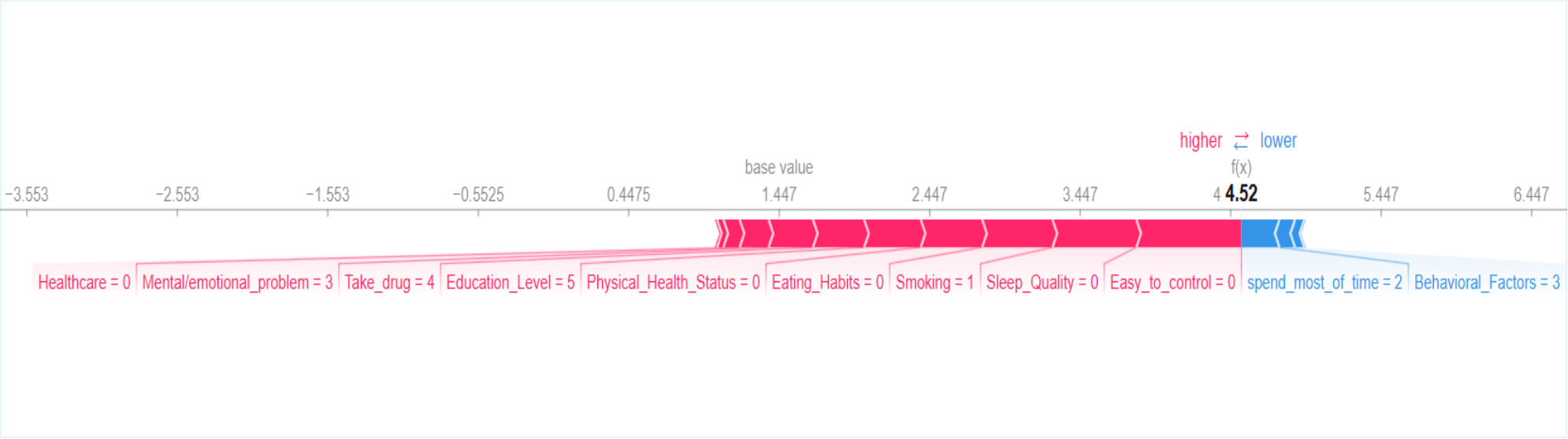}
    \caption{SHAP Force Plot Showing Feature Contributions for Sample 10.}
    \label{fig:SHAP2}
    \end{figure*}

Figure~\ref{fig:SHAP2} presents the SHAP force plot for sample 10, where the model output increases from $E[f(X)] = 1.447$ to $f(x) \approx 4.52$. Strong positive contributors include Easy\_to\_control = 0, Sleep\_Quality = 0, and Smoking = 1, supported by Eating\_Habits = 0, Physical\_Health\_Status = 0, Education\_Level = 5, and Take\_drug = 4, while spend\_most\_of\_time = 2 and Behavioral\_Factors = 3 slightly reduce the prediction. Overall, positive contributions dominate, driven by lifestyle, behavioral, and health-related factors leading to the high-risk prediction.

Previous studies on AI-based mental health prediction often relied on limited feature sets or single data modalities, reducing model generalizability \cite{Brito_10068843, Zhang_11167906, Acharya_14041630}. Many approaches struggled with class imbalance and small sample sizes, limiting predictive robustness \cite{chigagure2025machine, Zhang2025, Sekhar2024}. Our study improves accuracy and interpretability by using a multidimensional dataset, GAN-based oversampling, and ensemble feature selection with XAI.

\section{Conclusion and Future Work}
\label{sec:Con}
This paper presents an explainable multiclass mental health prediction model for drug-affected populations using hybrid PCA-IG feature selection, GAN-based oversampling, and DA-optimized XGBoost. The experimental results demonstrate superior performance, achieving high accuracy and outperforming traditional baseline models. The results highlight the prevailing influence of behavioral, lifestyle, and health-related variables, namely the quality of sleep, physical health, and emotional regulation, with the demographic characteristics playing a minor role in predictive results. The framework provided instance-level explainability with transparency, enhancing clinical trust, and facilitating informed decision-making through SHAP-based explainability. This study demonstrates that combining smart optimization, data balancing, and XAI improves the accuracy, reliability, and interpretability of predictive healthcare models.

The future studies will be aimed at testing the suggested framework on the bigger, multi-country, and more varied data to make it more practical and solid. The addition of multimodal and longitudinal data streams, including clinical data, neuroimaging data, and wearable sensor data, would be an additional step to predictive power and allow monitoring the progression of mental health in different drug-use contexts.

%\section*{Data Availability}  
%The data used in this study are available on Mendeley at the following link:
%\url{https://data.mendeley.com/datasets/b5ccpknz2z/1}

\bibliographystyle{ieeetr}
\bibliography{Ref}

@misc{Islam2024,
  author       = {Islam, Muksitul and Khan, Md. Forhad and Hasan Tusher, Mr. Raja Tariqul},
  title        = {Insights into Drug Addiction in Bangladesh: A Multidimensional Dataset},
  year         = {2024},
  publisher    = {Mendeley Data},
  version      = {V1},
  doi          = {10.17632/b5ccpknz2z.1}
}

@article{omuya2021feature1,
  title={Feature selection for classification using principal component analysis and information gain},
  author={Omuya, Erick Odhiambo and Okeyo, George Onyango and Kimwele, Michael Waema},
  journal={Expert Systems with Applications},
  volume={174},
  pages={114765},
  year={2021},
  publisher={Elsevier}
}

@inproceedings{ayon2024harvesting,
  title={Harvesting Insights: Unraveling Olive Dynamics and Climate Fluctuations Through Regression and SHapley Additive Explanations},
  author={Ayon, Shahriar Siddique and Hossain, Muhammad Ebrahim and Miah, Md Saef Ullah and Rahman, M Mostafizur and Mahmud, Mufti},
  booktitle={International Conference on Applied Intelligence and Informatics},
  pages={387--401},
  year={2024},
  organization={Springer}
}

@ARTICLE{Brito_10068843,
  author={Ximenes de Brito, Rhyan and Rolim Fernandes, Carlos Alexandre and Martins Moreira, Roberta Magda and Oliveira, Eliany Nazaré},
  journal={IEEE Latin America Transactions}, 
  title={Prediction Model for Common Mental Disorder and Depression in Users of Psychoactive Drugs}, 
  year={2023},
  volume={21},
  number={3},
  pages={399-407},
  doi={10.1109/TLA.2023.10068843}}

@article{Humayun2025,
  author       = {Humayun, Arsalan and Awang Nawi, Mohamad Arif Bin and Siddiqui, Muhammad Ilyas and Kabir, Russell and Babalola, Abdulhafeez},
  title        = {A Hybrid Mathematical Framework Combining Logistic Regression and Neural Networks with Explainable AI Techniques for Mental Health Prediction},
  journal      = {Contemporary Mathematics},
  year         = {2025},
  month        = {Sep},
  day          = {18},
  volume       = {6},
  number       = {5},
  pages        = {6521--6540},
  publisher    = {WiserPub},
  url          = {https://ojs.wiserpub.com/index.php/CM/article/view/8031}
}

@Article{Acharya_14041630,
AUTHOR = {Acharya, Nirmal and Kar, Padmaja and Ally, Mustafa and Soar, Jeffrey},
TITLE = {Predicting Co-Occurring Mental Health and Substance Use Disorders in Women: An Automated Machine Learning Approach},
JOURNAL = {Applied Sciences},
VOLUME = {14},
YEAR = {2024},
NUMBER = {4},
ARTICLE-NUMBER = {1630},
URL = {https://www.mdpi.com/2076-3417/14/4/1630},
ISSN = {2076-3417},
DOI = {10.3390/app14041630}
}

@article{chigagure2025machine,
  title={A machine learning approach for early prediction of mental health crises},
  author={Chigagure, Hassan and Sakala, Lucy Charity},
  journal={Computer Science and Information Technologies},
  volume={6},
  number={3},
  pages={335--345},
  year={2025}
}

@article{Squires2023,
  author       = {Squires, M. and Tao, X. and Elangovan, S. and Siddiqui, M. Ilyas and Kabir, Russell and Babalola, Abdulhafeez},
  title        = {Deep learning and machine learning in psychiatry: a survey of current progress in depression detection, diagnosis and treatment},
  journal      = {Brain Informatics},
  volume       = {10},
  number       = {10},
  year         = {2023},
  doi          = {10.1186/s40708-023-00188-6},
  url          = {https://doi.org/10.1186/s40708-023-00188-6}
}

@article{Zhang2025,
  author       = {Zhang, Zhen-Ze and Chen, Shao-Rong and Yu, Shen-Bao and Xia, Jie and Lin, Kai-Biao and Yang, Fan},
  title        = {MSFCL: Drug Combination Risk Level Prediction Based on Multi-Source Feature Fusion and Contrastive Learning},
  journal      = {Journal of Chemical Information and Modeling},
  year         = {2025},
  month        = {Jul},
  day          = {14},
  volume       = {65},
  number       = {13},
  pages        = {7285--7301},
  publisher    = {American Chemical Society},
  doi          = {10.1021/acs.jcim.5c00913},
  issn         = {1549-9596},
  url          = {https://doi.org/10.1021/acs.jcim.5c00913}
}

@INPROCEEDINGS{Yifan_10213032,
  author={Li, Yifan},
  booktitle={2023 3rd International Symposium on Computer Technology and Information Science (ISCTIS)}, 
  title={Application of Machine Learning to Predict Mental Health Disorders and Interpret Feature Importance}, 
  year={2023},
  volume={},
  number={},
  pages={257-261},
  doi={10.1109/ISCTIS58954.2023.10213032}}

@inproceedings{aziz2025comprehensive,
  title={A Comprehensive Framework Analysis of Cycle GAN-Based Modality Translation: Enhancing Brain Tumor Diagnostics from FLAIR to T2w},
  author={Aziz, Abidus Sattar and Sharif, Kazi Shaharair and Abubakkar, Md and Ahmad, Imran and Uddin, Mohammed Majbah},
  booktitle={2025 4th International Conference on Electronics Representation and Algorithm (ICERA)},
  pages={405--410},
  year={2025},
  organization={IEEE}
}

@article{ErnestOkonofua2025,
  author       = {Ernest-Okonofua, Excel Onajite and Ibadin, Franklin and Oyiborhoro, Ogheneyemarho Great and Mahmud, Aminu Sufu and Abengowe, Ifeoma Theodora and Akpohwaye, Peter Iroro},
  title        = {Impact of Unrestricted Drug Use on Psychiatric and Behavioral Disorders: Exploring Mental Health Effects in The United States},
  journal      = {International Journal of Multidisciplinary and Innovative Research},
  volume       = {2},
  number       = {3},
  pages        = {79--87},
  year         = {2025},
  month        = {Mar},
  doi          = {10.5281/zenodo.15004178},
  url          = {https://doi.org/10.5281/zenodo.15004178}
}

@Article{Zaha_13192543,
AUTHOR = {Zaha, Andreea Atena and Comșa, Antonia Lucia and Zaha, Dana Carmen and Vesa, Cosmin Mihai},
TITLE = {Patterns of Psychiatric Comorbidity Among Drug Users: A Prospective Observational Study in a Romanian Psychiatric Hospital},
JOURNAL = {Healthcare},
VOLUME = {13},
YEAR = {2025},
NUMBER = {19},
ARTICLE-NUMBER = {2543},
URL = {https://www.mdpi.com/2227-9032/13/19/2543},
PubMedID = {41095629},
ISSN = {2227-9032},
DOI = {10.3390/healthcare13192543}
}

@article{NARTENI2025110133,
title = {Explainable evaluation of generative adversarial networks for wearables data augmentation},
journal = {Engineering Applications of Artificial Intelligence},
volume = {145},
pages = {110133},
year = {2025},
issn = {0952-1976},
doi = {https://doi.org/10.1016/j.engappai.2025.110133},
url = {https://www.sciencedirect.com/science/article/pii/S0952197625001332},
author = {Sara Narteni and Vanessa Orani and Enrico Ferrari and Damiano Verda and Enrico Cambiaso and Maurizio Mongelli}
}

@article{Sekhar2024,
  author       = {Venkata Sekhar, D. and Purushotham Reddy, M. and Bhaswanth, N.},
  title        = {Feature Selection Based on Dragonfly Optimization for Psoriasis Classification},
  journal      = {International Journal of Intelligent Systems and Applications in Engineering},
  volume       = {12},
  number       = {3},
  pages        = {935--943},
  year         = {2024},
  month        = {Mar},
  url          = {https://ijisae.org/index.php/IJISAE/article/view/5374}
}

@INPROCEEDINGS{Zhang_11167906,
  author={Yi, Zhang},
  booktitle={2025 International Symposium on Intelligent Robotics and Systems (ISoIRS)}, 
  title={Constructing a Drug Consumption Prediction Model Based on Machine Learning Strategies}, 
  year={2025},
  volume={},
  number={},
  pages={1-6},
  doi={10.1109/ISoIRS65690.2025.11167906}}

@inproceedings{abubakkar2025explainable,
  title={Explainable suicide risk prediction with DeepFusion: A hybrid intelligence approach},
  author={Abubakkar, Md and Sharif, Kazi Shaharair and Ahmad, Imran and Tabila, Dill Mahzabina and Alsaud, Fowzy Alhussin and Debnath, Sajib},
  booktitle={2025 4th International Conference on Electronics Representation and Algorithm (ICERA)},
  pages={455--460},
  year={2025},
  organization={IEEE}
}

@misc{AIHW2025,
  author       = {{Australian Institute of Health and Welfare}},
  title        = {Mental illness and substance use},
  year         = {2025},
  publisher    = {Australian Institute of Health and Welfare},
  url          = {https://www.aihw.gov.au/mental-health/topic-areas/health-wellbeing/mental-illness-and-substance-use},
  note         = {Accessed: 2026-04-07}
}

@article{Erskine2015,
  author       = {Erskine, H. E. and Moffitt, T. E. and Copeland, W. E. and Costello, E. J. and Ferrari, A. J. and Patton, G. and Degenhardt, L. and Vos, T. and Whiteford, H. A. and Scott, J. G.},
  title        = {A heavy burden on young minds: the global burden of mental and substance use disorders in children and youth},
  journal      = {Psychological Medicine},
  volume       = {45},
  number       = {7},
  pages        = {1551--1563},
  year         = {2015},
  doi          = {10.1017/S0033291714002888}
}

@misc{PAHO2024,
  author       = {{Pan American Health Organization}},
  title        = {Over 3 million annual deaths due to alcohol and drug use, majority among men},
  year         = {2024},
  month        = {Jun},
  day          = {25},
  publisher    = {Pan American Health Organization},
  url          = {https://www.paho.org/en/news/25-6-2024-over-3-million-annual-deaths-due-alcohol-and-drug-use-majority-among-men},
  note         = {Accessed: 2026-04-07}
}

@article{Appleton2025,
  author       = {Appleton, R. and Barnett, P. and Clarke, C. and Yang, J. and Begum, S. and Edbrooke-Childs, J. and Emptage, I. and Foye, U. and Griffiths, J. L. and Hanson, I. and Hunt, N. C. and Jarvis, R. and McAuliffe, M. and Maynard, E. and Mitchell, L. and Mostafa, I. and Pemovska, T. and Saunders, R. and Trevillion, K. and Waite, P. and Lloyd-Evans, B. and Johnson, S.},
  title        = {Approaches to early intervention for common mental health problems in young people: a systematic review},
  journal      = {BMC Medicine},
  volume       = {23},
  number       = {1},
  pages        = {651},
  year         = {2025},
  month        = {Nov},
  day          = {24},
  doi          = {10.1186/s12916-025-04438-8},
  pmid         = {41286829},
  pmcid        = {PMC12642248},
  note         = {Impact Factor: 8.3, Q1}
}

@article{Toni2024,
  author       = {Toni, E. and Ayatollahi, H. and Abbaszadeh, R. and Fotuhi Siahpirani, A.},
  title        = {Machine Learning Techniques for Predicting Drug-Related Side Effects: A Scoping Review},
  journal      = {Pharmaceuticals (Basel)},
  volume       = {17},
  number       = {6},
  pages        = {795},
  year         = {2024},
  month        = {Jun},
  day          = {17},
  doi          = {10.3390/ph17060795},
  pmid         = {38931462},
  pmcid        = {PMC11206653},
  note         = {Impact Factor: 4.8, Q1}
}

@article{Ayon_10.1371,
    doi = {10.1371/journal.pone.0341168},
    author = {Ayon, Shahriar Siddique AND Al Mamun, Abdullah AND Hossain, Md. Ebrahim AND Alamro, Wasan AND Allawi, Yazan M. AND Prova, Nuzhat Noor Islam AND Miah, Md. Saef Ullah AND Sultan, Salman Md AND Abadleh, Ahmad},
    journal = {PLOS ONE},
    publisher = {Public Library of Science},
    title = {Explainable AI framework for improved Thalassemia mental health classification and feature selection},
    year = {2026},
    month = {01},
    volume = {21},
    url = {https://doi.org/10.1371/journal.pone.0341168},
    pages = {1-27},
    number = {1}

}

@article{Bari2026,
  author       = {Bari, S. and Vike, N. L. and Kim, B. W. and others},
  title        = {Predicting substance use behaviors with machine learning using small sets of judgment and contextual variables},
  journal      = {npj Mental Health Research},
  volume       = {5},
  pages        = {5},
  year         = {2026},
  doi          = {10.1038/s44184-025-00181-3},
  url          = {https://doi.org/10.1038/s44184-025-00181-3}
}

@article{Ndikumana2025,
  author       = {Ndikumana, F. and Izabayo, J. and Kalisa, J. and Nemerimana, M. and Nyabyenda, E. C. and Muzungu, S. H. and Komezusenge, I. and Uwase, M. and Ndagijimana, S. and Twizere, C. and Sezibera, V.},
  title        = {Machine learning-based predictive modelling of mental health in Rwandan Youth},
  journal      = {Scientific Reports},
  volume       = {15},
  number       = {1},
  pages        = {16032},
  year         = {2025},
  month        = {May},
  day          = {8},
  doi          = {10.1038/s41598-025-00519-z},
  pmid         = {40341215},
  pmcid        = {PMC12062285},
  note         = {Impact Factor: 3.9, Q1}
}

@misc{UNICEF2024,
  author       = {{United Nations Children's Fund (UNICEF)} and {{World Health Organization (WHO)}}},
  title        = {Increase in child and adolescent mental disorders spurs new push for action},
  year         = {2024},
  month        = {Jun},
  day          = {25},
  publisher    = {UNICEF and WHO},
  url          = {https://www.unicef.org/turkiye/en/press-releases/increase-child-and-adolescent-mental-disorders-spurs-new-push-action-unicef-and-who},
  note         = {Accessed: 2026-04-07}
}

@article{Tutun2023,
  author       = {Tutun, S. and Johnson, M. E. and Ahmed, A. and Albizri, A. and Irgil, S. and Yesilkaya, I. and Ucar, E. N. and Sengun, T. and Harfouche, A.},
  title        = {An AI-based Decision Support System for Predicting Mental Health Disorders},
  journal      = {Information Systems Frontiers},
  volume       = {25},
  number       = {3},
  pages        = {1261--1276},
  year         = {2023},
  doi          = {10.1007/s10796-022-10282-5},
  pmid         = {35669335},
  pmcid        = {PMC9142346},
  note         = {Impact Factor: 8.3, Q1. Epub 2022 May 28}
}

@inproceedings{Ayon_Advancing_2025,
  author       = {Ayon, S. S. and Hossain, M. E. and Miah, M. S. U. and Rahman, M. M. and Mahmud, M.},
  title        = {Advancing Mental Health Problems with Machine Learning and Genetic Algorithms for Anxiety Classification in Bangladeshi University Students},
  booktitle    = {Brain Informatics. BI 2024},
  editor       = {Itthipuripat, S. and Ascoli, G. A. and Li, A. and Pat, N. and Kuai, H.},
  series       = {Lecture Notes in Computer Science},
  volume       = {15541},
  publisher    = {Springer},
  address      = {Singapore},
  year         = {2025},
  doi          = {10.1007/978-981-96-3294-7_26},
  url          = {https://doi.org/10.1007/978-981-96-3294-7_26}
}
\end{document}